\documentclass[11pt,a4paper]{article}
\usepackage[hyperref]{emnlp-ijcnlp-2019}
\usepackage{times}
\usepackage{latexsym}

\usepackage{url}

\usepackage{amsmath}
\usepackage{graphicx}
\usepackage{caption}
\usepackage{subfigure}
\usepackage{amsfonts}

\usepackage{multirow}
\usepackage{array}

\makeatletter
\def\hlinew#1{%
    \noalign{\ifnum0=`}\fi\hrule \@height #1 \futurelet
    \reserved@a\@xhline}
\makeatother

\aclfinalcopy 

\title{Generating Questions for Knowledge Bases \\via Incorporating Diversified Contexts and Answer-Aware Loss}

\author{
    Cao Liu$^{1,2}$,
    Kang Liu$^{1,2}$,
    Shizhu He$^{1,2}$,
    Zaiqing Nie$^3$, 
    Jun Zhao$^{1,2}$ \\
    $^1$ National Laboratory of Pattern Recognition, Institute of Automation, \\
    Chinese Academy of Sciences, Beijing, 100190, China \\
    $^2$ University of Chinese Academy of Sciences, Beijing, 100049, China \\
    $^3$ Alibaba AI Labs, Beijing, 100029, China \\
    \{cao.liu, kliu, shizhu.he, jzhao\}@nlpr.ia.ac.cn \\
    zaiqing.nzq@alibaba-inc.com
}

\date{}

\begin{document}
\maketitle
\begin{abstract}
    We tackle the task of question generation over knowledge bases. Conventional methods for this task neglect two crucial research issues: 1) the given predicate needs to be expressed; 2) the answer to the generated question needs to be definitive. In this paper, we strive toward the above two issues via incorporating diversified contexts and answer-aware loss. Specifically, we propose a neural encoder-decoder model with multi-level copy mechanisms to generate such questions. Furthermore, the answer aware loss is introduced to make generated questions corresponding to more definitive answers. Experiments demonstrate that our model achieves state-of-the-art performance. Meanwhile, such generated question can express the given predicate and correspond to a definitive answer. 
\end{abstract}

\section{Introduction}
\label{Setion: Introduction}
Question Generation over Knowledge Bases (KBQG) aims at generating natural language questions for the corresponding facts on KBs, and it can benefit some real applications. Firstly, KBQG can automatically annotate question answering (QA) datasets. Secondly, the generated questions and answers will be able to augment the training data for QA systems. More importantly, KBQG can improve the ability of machines to actively ask questions on human-machine conversations \cite{D17-1090,D18-1427}. Therefore, this task has attracted more attention in recent years \cite{P16-1056,N18-1020}.

\begin{figure}[t]
    \begin{center}
        \includegraphics[width=218pt]{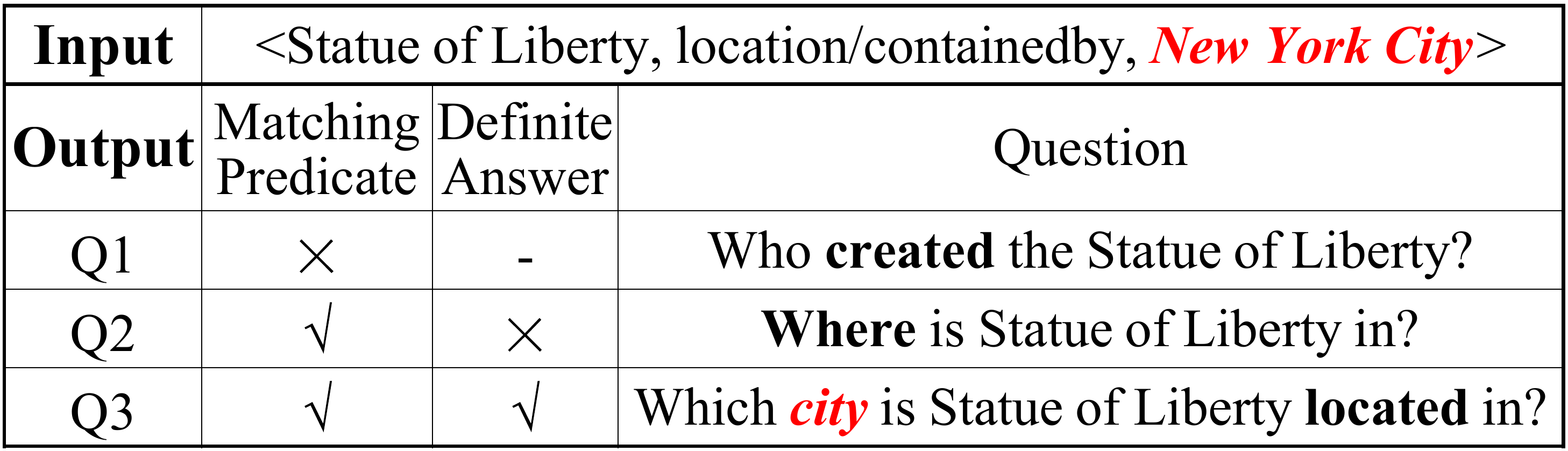}
        \caption{Examples of KBQG. We aims at generating questions like Q3 which expresses (matches) the given predicate and refers to a definitive answer.}
        \label{fig:example}
    \end{center}
\end{figure}

Specifically, KBQG is the task of generating natural language questions according to the input facts from a knowledge base with triplet form, like $<$subject, predicate, object$>$. For example, as illustrated in Figure \ref{fig:example}, KBQG aims at generating a question ``Which city is Statue of Liberty located in?" (Q3) for the input factual triplet ``$<$Statue of Liberty, location/containedby\footnote{We omit the domain of the predicate for sake of brevity.\label{omit_domain}}, New York City$>$". Here, the generated question is associated to the subject ``\textit{Statue of Liberty}" and the predicate \texttt{fb:location/containedby}) of the input fact, and the answer corresponds to the object ``\textit{New York City}".

As depicted by \newcite{P16-1056}, KBQG is required to transduce the triplet fact into a question about the subject and predicate, where the object is the correct answer. Therefore, it is a key issue for KBQG to correctly understand the knowledge symbols (subject, predicate and object in the triplet fact) and then generate corresponding text descriptions. More recently, some researches have striven toward this task, where the behind intuition is to construct implicit associations between facts and texts. Specifically, \newcite{P16-1056} designed an encoder-decoder architecture to generate questions from structured triplet facts. In order to improve the generalization for KBQG, \newcite{N18-1020} utilized extra contexts as input via distant supervisions \cite{Mintz:2009:DSR:1690219.1690287}, then a decoder is equipped with attention and part-of-speech (POS) copy mechanism to generate questions. Finally, this model obtained significant improvements. Nevertheless, we observe that there are still two important research issues (RIs) which are not processed well or even neglected. 

\textit{\textbf{RI-1:} The generated question is required to \textbf{express the given predicate} in the fact}. For example in Figure \ref{fig:example}, Q1 does not express (match) the predicate (\texttt{fb:location/containedby}) while it is expressed in Q2 and Q3. Previous work \cite{N18-1020} usually obtained predicate textual contexts through distant supervision. However, the distant supervision is noisy or even wrong (e.g. ``X is the husband of Y" is the relational pattern for the predicate \texttt{fb:marriage/spouse}, so it is wrong when ``X" is a woman). Furthermore, many predicates in the KB have no predicate contexts. We make statistic in the resources released by \newcite{N18-1020}, and find that only 44\% predicates have predicate textual context\footnote{We map the ``prop\_text\_evidence.csv" file to the ``property.vocab" file in \newcite{N18-1020}\label{predicate context percentage}}. Therefore, it is prone to generate error questions from such without-context predicates. 

\textit{\textbf{RI-2:} The generated question is required to contain a \textbf{definitive answer}}. A definitive answer means that one question only associates with a determinate answer rather than alternative answers. As an example in Figure \ref{fig:example}, Q2 may contain ambiguous answers since it does not express the refined answer type. As a result, different answers including ``\textit{United State}", ``\textit{New York City}", etc. may be correct. In contrast, Q3 refers to a definitive answer (the object ``\textit{New York City}" in the given fact) by restraining the answer type to a city. We believe that Q3, which expresses the given predicate and refers to a definitive answer, is a better question than Q1 and Q2. In previous work, \newcite{N18-1020} only regarded a most frequently mentioned entity type as the textual context for the subject or object in the triplet. In fact, most answer entities have multiple types, where the most frequently mentioned type tends to be universal (e.g. a broad type ``administrative region" rather than a refined type ``US state" for the entity ``\textit{New York}"). Therefore, generated questions from \newcite{N18-1020} may be difficult to contain definitive answers.

To address the aforementioned two issues, we exploit more \textbf{diversified contexts} for the given facts as textual contexts in an encoder-decoder model. Specifically, besides using predicate contexts from the distant supervision utilized by \newcite{N18-1020}, we further leverage the domain, range and even topic for the given predicate as contexts, which are off-the-shelf in KBs (e.g. the range and the topic for the predicate \texttt{fb:location/containedby} are ``location" and ``containedby", respectively\textsuperscript{\ref{omit_domain}}). Therefore, 100\% predicates (rather than 44\%\textsuperscript{\ref{predicate context percentage}} of those in \citeauthor{N18-1020}) have contexts. Furthermore, in addition to the most frequently mentioned entity type as contexts used by \newcite{N18-1020}, we leverage the type that best describes the entity as contexts (e.g. a refined entity type\footnote{We obtain such representative entity types through the predicate \texttt{fb:topic/notable\_types} in freebase.\label{note:notable_type}} ``US state" combines a broad type ``administrative region" for the entity ``New York"), which is helpful to refine the entity information. Finally, in order to make full use of these contexts, we propose context-augmented fact encoder and multi-level copy mechanism (KB copy and context copy) to integrate diversified contexts, where the multi-level copy mechanism can copy from KB and textual contexts simultaneously. For the purpose of further making generated questions correspond to definitive answers, we propose the \textbf{answer-aware loss} by optimizing the cross-entropy between the generated question and answer type words, which is beneficial to generate precise questions. 

We conduct experiments on an open public dataset. Experimental results demonstrate that the proposed model using diversified textual contexts outperforms strong baselines (+4.5 BLEU4 score). Besides, it can further increase the BLEU score (+5.16 BLEU4 score) and produce questions associated with more definitive answers by incorporating answer-aware loss. Human evaluations complement that our model can express the given predicate more precisely. 

In brief, our main contributions are as follows:

(1) We leverage diversified contexts and multi-level copy mechanism to alleviate the issue of incorrect predicate expression in traditional methods.

(2) We propose an answer-aware loss to tackle the issue that conventional methods can not generate questions with definitive answers.

(3) Experiments demonstrate that our model achieves state-of-the-art performance. Meanwhile, such generated question can express the given predicate and refer to a definitive answer.

\begin{figure*}[t]
    \begin{center}
        \includegraphics[width=446pt]{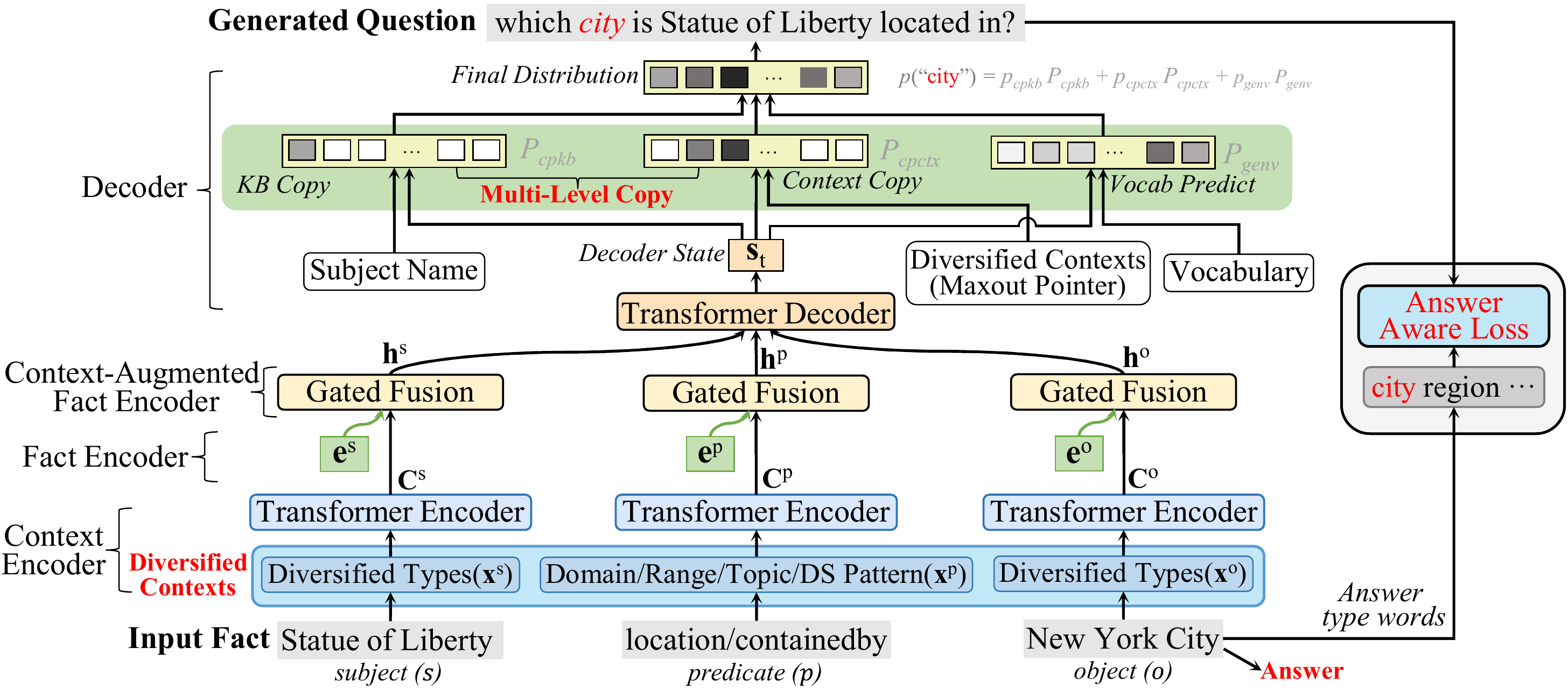}
        \caption{Overall structure of the proposed model for KBQG. A \textbf{context encoder} is firstly employed to encode each textual context (Sec. \ref{subsec:ctx_enc}), where ``Diversified Types" represents the subject (object) context, and ``DS pattern" denotes the relational pattern from distant supervisions. At the same time, a \textbf{fact encoder} transforms the fact into low-dimensional representations (Sec. \ref{subsec:fact_enc}). The above two encoders are aggregated by the \textbf{context-augmented fact encoder} (Sec. \ref{subsec:ctx_fact_enc}). Finally, the aggregated representations are fed to the \textbf{decoder} (Sec. \ref{subsec:dec}), where the decoder leverages multi-level copy mechanism (KB copy and context copy) to generate target question words.}
        \label{fig:overall-structure}
    \end{center}
\end{figure*}

\section{Task Description}
We leverage textual contexts concerned with the triplet fact to generate questions over KBs. The task of KBQG can be formalized as follows:
\begin{equation}
\begin{split}
P(Y|F)=\prod \nolimits_{t=1}^{|Y|}P(y_{t}|y_{<t},F, C)
\end{split}
\label{eq:3411}
\end{equation}
where $F=(s,p,o)$ represents the subject ($s$), predicate ($p$) and object ($o$) of the input triplet, $C=\{\textbf{x}^{s}, \textbf{x}^{p}, \textbf{x}^{o}\}$ denotes a set of additional textual contexts, $Y=(y_1,y_2,...,y_{|Y|})$ is the generated question, $y_{<t}$ represents all previously generated question words before time step $t$. 

\section{Methodology}

Our model extends the encoder-decoder architecture \cite{cho-EtAl:2014:EMNLP2014} with three encoding modules and two copy mechanisms in the decoder. The model overview is shown in Figure \ref{fig:overall-structure} along with its caption. It should be emphasized that we additionally design an answer-aware loss to make the generated question associated with a definitive answer (Sec. \ref{subsubsec:ans_loss}). 

\subsection{Context Encoder}
\label{subsec:ctx_enc}
Inspired by the great success of transformer \cite{NIPS2017_7181} in sequence modeling \cite{shen2018disan}, we adopt a transformer encoder to encode each textual context separately. 
Take the subject context $\textbf{x}^{s}$ as an example, $\textbf{x}^{s}=(x_{1}^{s}, x_{2}^{s},...,x_{|s|}^{s})$ is concatenated from diversified types for the subject, and $x_{i}^{s}$ is the $i$-th token in the subject context, $|s|$ stands for the length of the subject context. Firstly, $\textbf{x}^{s}$ is mapped into a query matrix $\textbf{Q}$, where $\textbf{Q}$ is constructed by summing the corresponding token embeddings and segment embeddings. Similar to BERT \cite{cotterell-eisner-2018-deep}, segment embeddings are the same for tokens of $\textbf{x}^{s}$ but different for that of $\textbf{x}^{p}$ (predicate context) or $\textbf{x}^{o}$ (object context). Based on the query matrix, transformer encoder works as follows:
\begin{gather}\footnotesize
    \textbf{Q}_{j}=\textbf{QW}_{j}^{Q}, \textbf{K}_{j}=\textbf{KW}_{j}^{K}, \textbf{V}_{j}=\textbf{VW}_{j}^{V} \label{eq:subspace} \\
    head_{j} = {\rm softmax} (\textbf{Q}_{j}\textbf{K}_{j}^{T} / \sqrt{d/h})\textbf{V}_{j} \label{eq:head} \\
    \textbf{H}^{s}={\rm Concat}(head_1,head_2,...,head_{h})\textbf{W}_{0} \label{eq:concat} \\
    \textbf{N}^{s} = {\rm LayerNorm}(\textbf{Q} + \textbf{H}^{s}) \label{eq:layernorm} \\
    \textbf{C}^{s} = {\rm max}(0, \textbf{N}^{s} \textbf{W}_{1} + b_{1}) \textbf{W}_{2} + b_{2} \label{eq:FFN}
\end{gather}     
where $\textbf{K}$ and $\textbf{V}$ are the key matrix and value matrix, respectively. It is called self-attention because $\textbf{K}$ and $\textbf{V}$ are equal to the query matrix $\textbf{Q} \in \mathbb{R} ^{|s|, d}$ in the encoding stage, where $d$ represents the number of hidden units. And $h$ denotes the number of the heads in multi-head attention mechanism of the transformer encoder. It first projects the input matrixes ($\textbf{Q}$, $\textbf{K}$, $\textbf{V}$) into subspaces $h$ times mapped by different linear projections $\textbf{W}_{j}^{Q}$, $\textbf{W}_{j}^{K}$, $\textbf{W}_{j}^{V} \in \mathbb{R} ^{|s|, d/h}$ ($j=1,2,...,h$) in Equation \ref{eq:subspace}. And then $h$ projections perform the scaled dot-product attention to obtain the representation of each head in parallel (Equation \ref{eq:head}). Representations for all parallel heads are concatenated together in Equation \ref{eq:concat}. After residual connection, layer normalization (Equation \ref{eq:layernorm}) and feed forward operation (Equation \ref{eq:FFN}), we can obtain the subject context matrix $\textbf{C}^{s}=\{\textbf{c}_{1}^{s},\textbf{c}_{2}^{s},...,\textbf{c}_{|s|}^{s}\} \in \mathbb{R} ^{|s|, d}$.

Similarly, $\textbf{C}^{p}$ and $\textbf{C}^{o}$ are obtained from the same transformer encoder for the predicate and object, respectively.

\subsection{Fact Encoder}
\label{subsec:fact_enc}
In contrast to general Sequence-to-Sequence (Seq2Seq) model \cite{sutskever2014sequence}, the input fact is not a word sequence but instead a structured triplet $F=(s,p,o)$. We employ a fact encoder to transform each atom in the fact into a fixed embedding, and the embedding is obtained from a KB embedding matrix. For example, the subject embedding $\textbf{e}^{s} \in \mathbb{R} ^{d}$ is looked up from the KB embedding matrix $\textbf{E}_{f} \in \mathbb{R}^{k,d}$, where $k$ represents the size of KB vocabulary, and the size of KB embedding is equal to the number of hidden units ($d$) in Equation \ref{eq:head}. Similarly, the predicate embedding $\textbf{e}^{p}$ and the object embedding $\textbf{e}^{o}$ are mapped from the KB embedding matrix $\textbf{E}_{f}$, where $\textbf{E}_{f}$ is pre-trained using \textit{TransE} \cite{NIPS2013_5071} to capture much more fact information in previous work \cite{N18-1020}. In our model, $\textbf{E}_{f}$ can be pre-trained or randomly initiated (Details in Sec. \ref{subsec:Pre-trained KB}).

\subsection{Context-Augmented Fact Encoder}
\label{subsec:ctx_fact_enc}
In order to combine both the context encoder information and the fact encoder information, we propose a context-augmented fact encoder which applies the gated fusion unit \cite{W18-2601} to integrate the context matrix and the fact embedding. For example, the subject context matrix $\textbf{C}^{s}=\{\textbf{c}_{1}^{s},\textbf{c}_{2}^{s},...,\textbf{c}_{|s|}^{s}\}$ and the subject embedding vector $\textbf{e}^{s}$ are integrated by the following gated fusion:
\begin{gather}
    \textbf{f} = {\rm tanh}(\textbf{W}_f[\textbf{c}^{s},\textbf{e}^{s}]) \label{eq:f} \\
    \textbf{g} = {\rm sigmoid}(\textbf{W}_g[\textbf{c}^{s},\textbf{e}^{s}]) \label{eq:g} \\
    \textbf{h}^{s} = \textbf{g} \ \odot \ \textbf{f} + (1-\textbf{g})  \ \odot \ \textbf{e}^{s} \label{eq:fusion}
\end{gather}
where $\textbf{c}^{s}$ is an attentive vector from $\textbf{e}^{s}$ to $\textbf{C}^{s}$, which is similar to \newcite{D18-1424}. The attentive vector $\textbf{c}^{s}$ is combined with original subject embedding $\textbf{e}^{s}$ as a new enhanced representation $\textbf{f}$ (Equation \ref{eq:f}). And then a learnable gate vector, $\textbf{g}$ (Equation \ref{eq:g}), controls the information from $\textbf{c}^{s}$ and $\textbf{e}^{s}$ to the final augmented subject vector $\textbf{h}^{s} \in \mathbb{R} ^{d}$ (Equation \ref{eq:fusion}), where $\odot$ denotes the element-wise multiplication. Similarly, the augmented predicate vector $\textbf{h}^{p}$ and the augmented object vector $\textbf{h}^{o}$ are calculated in the same way. Finally, the context-augmented fact representation $\textbf{H}_{f} \in \mathbb{R} ^{3,d}$ is the concatenation of augmented vectors as follows:\begin{equation}
\begin{split}
\textbf{H}_{f} = [\textbf{h}^{s}; \textbf{h}^{p}; \textbf{h}^{o}]
\end{split}
\label{eq:3411}
\end{equation}

\subsection{Decoder}
\label{subsec:dec}
The decoder aims at generating a question word sequence. As shown in Figure \ref{fig:overall-structure}, we also exploit the transformer as the basic block in our decoder. Then we use a multi-level copy mechanism (KB copy and context copy), which allows copying from KBs and textual contexts.

Specifically, we first map the input of the decoder into an embedding representation by looking up word embedding matrix, then we use position embedding \cite{NIPS2017_7181} to enhance sequential information. Compared with the transformer encoder in Sec. \ref{subsec:ctx_enc}, transformer decoder has an extra sub-layer: a fact multi-head attention layer, which is similar to Equation \ref{eq:subspace}-\ref{eq:FFN}, where the query matrix is initiated with previous decoder sub-layer while both the key matrix and the value matrix are the augmented fact representation $\textbf{H}_{f}$. After feedforward and multiple transformer layers, we obtain the decoder state $\textbf{s}_t$ at time step $t$, and then $\textbf{s}_t$ could be leveraged to generate the target question sequence word by word.

As depicted in Figure \ref{fig:overall-structure}, we propose multi-level copy mechanism to generate question words. At each time step $t$, given decoder state $\textbf{s}_t$ together with input fact $F$, textual contexts $C$ and vocabulary $V$, the probabilistic function for generating any target question word $y_t$ is calculated as: 
\begin{gather}
    \begin{split}
        P(y_{t}|\textbf{s}_t, \! y_{t-1}, \! F, \! C) \! = \! p_{genv} P_{genv}(y_{t}|\textbf{s}_t, \! V) +\!\\
        p_{cpkb} P_{cpkb}(y_{t}|\textbf{s}_t, \!F) \! + \! p_{cpctx} P_{cpctx}(y_{t}|\textbf{s}_t,\! C) \label{eq:yt}
    \end{split} \\
    p_{genv}, p_{cpkb}, p_{cpctx} \! = \! {\rm softmax}([\textbf{s}_t, \textbf{y}_{t-1}]) \label{eq:mode}
\end{gather}
where $genv$, $cpkb$ and $cpctx$ denote the vocab generation mode, the KB copy mode and the context copy mode, respectively. In order to control the balance among different modes, we employ a 3-dimensional switch probability in Equation \ref{eq:mode}, where $\textbf{y}_{t-1}$ is the embedding of previous generated word, $P_{\cdot}(\cdot | \cdot)$ indicates the probabilistic score function for generated target word of each mode. In the three probability score functions, $P_{vocab}$ is typically performed by a $softmax$ classifier over a fixed vocabulary $V$ based on the word embedding similarity, and the details of $P_{cpkb}$ and $P_{cpctx}$ are in the following.
\subsubsection{KB Copy}
Previous study found that most questions contain the subject name or its aligns in SimpleQuestion \cite{D18-1051}. However, the predicate name and object name hardly appear in the question. Therefore, we only copy the subject name in the KB copy, where $P_{cpkb}(y_{t}|\textbf{s}_t,f)$, the probability of copying the subject name, is calculated by a neural network function with a multi-layer perceptron (MLP) projected from $\textbf{s}_t$.

\subsubsection{Context Copy}
\newcite{N18-1020} demonstrated the effectiveness of POS copy for the context. However, such a copy mechanism heavily relies on POS tagging. Inspired by the CopyNet \cite{P16-1154}, we directly copy words in the textual contexts $C$, and it does not rely on any POS tagging. Specifically, the input sequence $\chi$ for the context copy is the concatenation of all words in the textual contexts $C$. Unfortunately, $\chi$ is prone to contain repeated words because it consists of rich contexts for subject, predicate and object. The repeated words in the input sequence tend to cause repetition problems in output sequences \cite{P16-1008}. We adopt the maxout pointer \cite{D18-1424} to address the repetition problem. Instead of summing all the probabilistic scores for repeated input words, we limit the probabilistic score of repeated words to their maximum score as Equation \ref{eq:cpctx}:
\begin{equation}
\begin{split}
P_{cpctx}(y_t|.)\!\!=\!\!\left\{\!\begin{matrix}
\underset{m,\, where\, \chi_{m}=y_{t}}{{\rm max}} sc_{t,m} \quad  y_{t} \! \in \! \chi \\ 
~~~~~~~~~~0 ~~~~~~~~~~~~~~~ otherwise
\end{matrix}\right. 
\end{split}
\label{eq:cpctx}
\end{equation}
where $\chi_{m}$ represents the $m$-th token in the input context sequence $\chi$, $sc_{t,m}$ is the probabilistic score of generating the token $\chi_m$ at time step $t$, and $sc_{t,m}$ is calculated by a softmax function over $\chi$.

\subsection{Learning}
\subsubsection{Question-Aware Loss}
\label{subsubsec:ques_loss}
It is totally differential for our model to obtain question words, and it can be optimized in an end-to-end manner by back-propagation. Given the input fact $F$, additional textual context $C$ and target question word sequence $Y$, the object function is to optimize the following negative log-likelihood:\begin{equation}
\begin{split}
\mathcal{L}_{ques\_loss}\!=\!\frac{-1}{|Y|}\! \sum _{t=1}^{|Y|} \! {\rm log}[P(y_t|\textbf{s}_t,\!y_{t-1},\!F,\!C)]
\end{split}
\label{eq:ques_loss}
\end{equation}

The question-aware loss $\mathcal{L}_{ques\_loss}$ does not require any additional labels to optimize because the three modes share a same ${\rm softmax}$ classifier to keep a balance (Equation \ref{eq:mode}), and they can learn to coordinate each other by minimizing $\mathcal{L}_{ques\_loss}$.

\subsubsection{Answer-Aware loss}
\label{subsubsec:ans_loss}
It is able to generate questions similar to the labeled questions by optimizing the question-aware loss $\mathcal{L}_{ques\_loss}$. However, there is an ambiguous problem in the annotated questions where the questions have alternative answers rather than determinate answers \cite{D18-1051}. In order to make generated questions correspond to definitive answers, we propose a novel answer-aware loss. By answer-aware loss, we aim at generating an answer type word in the question, which contributes to generating a question word matching the answer type. Formally, the answer-aware loss is in the following: 
\begin{equation}
\begin{split}
\mathcal{L}_{ans\_loss}=\underset{a_n,a_n \in A}{{\rm min}} \ \ \underset{y_t,y_t \in Y}{{\rm min}} \ H_{a_n,y_t}
\end{split}
\label{eq:ans_loss}
\end{equation}
where $A=\{a_n\}_{n=1}^{|A|}$ is a set of answer type words. We treat object type words as the answer type words because the object is the answer. $H_{a_n,y_t}$ denotes the cross entropy between the answer type word $a_n$ and the generated question word $y_t$. Finally, the minimum cross entropy is regarded as the answer-aware loss $\mathcal{L}_{ans\_loss}$. Optimizing $\mathcal{L}_{ans\_loss}$ means that the model aims at generating an answer type word in the generated question sequence. For example, the model tends to generate Q3 rather than Q2 in Figure \ref{fig:example}, because Q3 contains an answer type word---``city". Similarly, $\mathcal{L}_{ans\_loss}$ could be optimized in an end-to-end manner, and it can integrate $\mathcal{L}_{ques\_loss}$ by a weight coefficient $\lambda$ to the total loss as follows:
\begin{equation}
\begin{split}
\mathcal{L}_{total\_loss}=\mathcal{L}_{ques\_loss} + \lambda \mathcal{L}_{ans\_loss}
\end{split}
\label{eq:toal_loss}
\end{equation}

\section{Experiment}
\label{Setion: Experiment}
\subsection{Experimental Settings}
\label{subsec:exp_setting}


\subsubsection{Experimental Data Details}
We conduct experiments on the SimpleQuestion dataset \cite{DBLP:journals/corr/BordesUCW15}, and there are 75910/10845/21687 question answering pairs (QA-pairs) for training/validation/test. In order to obtain \textbf{diversified contexts}, we additionally employ domain, range and topic of the predicate to improve the coverage of predicate contexts. In this way, 100\% predicates (rather than 44\%\textsuperscript{\ref{predicate context percentage}} of those in \citeauthor{N18-1020}) have contexts. For the subject and object context, we combine the most frequently mentioned entity type \cite{N18-1020} with the type that best describe the entity\textsuperscript{\ref{note:notable_type}}. The KB copy needs subject names as the copy source, and we map entities with their names similar to those in \newcite{N18-2047}. The data details are in Appendix A and submitted Supplementary Data. 

\subsubsection{Evaluation Metrics}
\label{subsubsec:metrics}
Following \cite{P16-1056,N18-1020}, we adopt some word-overlap based metrics (WBMs) for natural language generation including BLEU-4 \cite{Papineni:2002:BMA:1073083.1073135}, ROUGE$_{\rm L}$ \cite{Lin:2004} and METEOR \cite{W14-3348}. However, such metrics still suffer from some limitations \cite{novikova-EtAl:2017:EMNLP2017}. Crucially, it might be difficult for them to measure whether generated questions that express the given predicate and refer to definitive answers. To better evaluate generated questions, we run two further evaluations as follows.

(1) \textbf{\textit{Predicate identification}}: Following \newcite{N18-2047}, we employ annotators to judge whether the generated question expresses the given predicate in the fact or not. The score for predicate identification is the percentage of generated questions that express the given predicate. 

(2) \textbf{\textit{Answer coverage}}: We define a novel metric called answer coverage to identify whether the generated question refers to a definitive answer. Specifically, answer coverage is obtained by automatically calculating the percentage of questions that contain answer type words, and answer type words are object contexts (entity types for the object are regarded as answer type words).

Furthermore, it is hard to automatically evaluate the naturalness of generated questions. Following \newcite{N18-2047}, we adopt human evaluation to measure the naturalness by a score of 0-5.

\subsubsection{Comparison with State-of-the-arts}
We compare our model with following methods.

(1) \textit{Template}: A baseline in \newcite{P16-1056}, it randomly chooses a candidate fact $F_c$ in the training data to generate the question, where $F_c$ shares the same predicate with the input fact.

(2) \textit{\newcite{P16-1056}}: We compare our methods with the single placeholder model, which performs best in \newcite{P16-1056}.

(3) \textit{\newcite{N18-1020}}: We compare our methods with the model utilizing copy actions, the best performing model in \newcite{N18-1020}. Although this model is designed to a zero-shot setting (for unseen predicates and entity type), it has good abilities to generate better questions (on known or unknown predicates and entity types) represented in the additional context input and SPO copy mechanism.

\subsubsection{Implementation Details}
To make our model comparable to the comparison methods, we keep most parameter values the same as \newcite{N18-1020}. We utilize RMSProp algorithm with a decreasing learning rate (0.001), batch size (200) to optimize the model. The size of KB embeddings is 200, and KB embeddings are pre-trained by TransE \cite{NIPS2013_5071}. The word embeddings are initialized by the pre-trained Glove word vectors\footnote{http://nlp.stanford.edu/data/glove.6B.zip} with 200 dimensions. In the transformer, we set the hidden units $d$ to 200, and we employ 4 paralleled attention head and a stack of 5 identical layers. We set the weight ($\lambda$) of the answer-aware loss to 0.2.

\subsection{Overall Comparisons}
\label{subsec:Overall Comparisons}
\begin{table}[h]\footnotesize
    \centering
    \setlength{\tabcolsep}{4pt}
    \renewcommand\arraystretch{1.1}
    \begin{tabular}{p{2.6cm}<{\centering}p{1.2cm}<{\centering}p{1.3cm}<{\centering}p{1.3cm}<{\centering}}
        \hlinew{1.2pt}
        Model $\quad \ \ $ & BLEU4 & ROUGE$\rm _{L}$ & METEOR \tabularnewline
        \hlinew{1.2pt}
        Template $\quad \ \ $ & 31.36 & * & 33.12 \tabularnewline
        \newcite{P16-1056} & 33.32 & * & 35.38 \tabularnewline
        \newcite{N18-1020} & 36.56 & 58.09 & 34.41 \tabularnewline
        \hline
        Our Model $\quad \ \ \,$ & 41.09 & 68.68 & 47.75 \tabularnewline
        $\ \ \,$ Our Model$_{\rm ans\_loss}$ & \textbf{41.72} & \textbf{69.31} & \textbf{48.13} \tabularnewline
        \hlinew{1.2pt}
    \end{tabular}
    \caption{Overall comparisons on the test data, where ``ans\_loss" represents answer-aware loss.}
    \label{tab: overall performance}
\end{table}
In Table \ref{tab: overall performance}, we compare our model with the typical baselines on word-overlap based metrics. It is evident that our model is remarkably better than baselines on all metrics, where the BLEU4 score increases 4.53 compared with the strongest baseline \cite{N18-1020}. Especially, incorporating answer-aware loss (the last line in Table \ref{tab: overall performance}) further improves the performance (+5.16 BLEU4).

\subsection{Performances on Predicate Identification}
\label{subsec:Predicate Identification}
\begin{table}[h]\footnotesize
    \centering
    \setlength{\tabcolsep}{4pt}
    \renewcommand\arraystretch{1.1}
    \begin{tabular}{p{2.8cm}<{\centering}p{2.5cm}<{\centering}}
        \hlinew{1.2pt}
        Model & Pred. Identification \tabularnewline
        \hlinew{1.2pt}
        \newcite{P16-1056} & 53.5 \tabularnewline
        \newcite{N18-1020} & 71.5 \tabularnewline
        Our Model$_{\rm ans\_loss}$ & \textbf{75.5} \tabularnewline
        \hlinew{1.2pt}
    \end{tabular}
    \caption{Performances on predicate identification.}
    \label{tab: pred_identification}
\end{table}
To evaluate the ability of our model on predicate identification, we sample 100 generated questions from each model, and then two annotators are employed to judge whether the generated question expresses the given predicate. The Kappa for inter-annotator statistics is 0.611, and p-value for all scores is less than 0.005. As shown in Table \ref{tab: pred_identification}, we can see that our model has a significant improvement in the predicate identification.

\subsection{Performances on Answer Coverage --- The Effectiveness of Answer-Aware Loss}
\label{subsec:Answer Coverage}
\begin{table}[h]\footnotesize
    \centering
    \setlength{\tabcolsep}{4pt}
    \renewcommand\arraystretch{1.1}
    \begin{tabular}{p{2.6cm}<{\centering}p{1.0cm}<{\centering}p{1.2cm}<{\centering}p{1.3cm}<{\centering}}
        \hlinew{1.2pt}
        Model & $\lambda$ & BLEU4 & ${\rm Ans_{cov}}$ \tabularnewline
        \hlinew{1.2pt}
        \newcite{N18-1020} & 0 & 36.56 & 59.49 \tabularnewline
        Our Model& 0 & 41.09 & 61.65 \tabularnewline \hline
        Our Model$_{\rm ans\_loss}$ & 0.05 & 41.55 & 62.27 \tabularnewline
        Our Model$_{\rm ans\_loss}$ & 0.2 & \textbf{41.72} & 64.23 \tabularnewline
        Our Model$_{\rm ans\_loss}$ & 0.5 & 41.57 & \textbf{65.50} \tabularnewline
        Our Model$_{\rm ans\_loss}$ & 1.0 & 41.34 & 65.25 \tabularnewline
        \hlinew{1.2pt}
    \end{tabular}
    \caption{Performances on answer coverage, where ``Ans$_{\rm cov}$" denotes the metric of answer coverage. ``$\lambda$" is the weight of the answer-aware loss in Equation \ref{eq:toal_loss}.}
    \label{tab: answer-aware loss}
\end{table}
Table \ref{tab: answer-aware loss} reports performances on BLUE4 and answer coverage (Ans$_{\rm cov}$). We can obtain that: 

(1) When answer-aware loss is not leveraged ($\lambda=0$), advantages of performance are obvious in our model. Note that the answer coverage is 55.23 on the human-labeled questions. Although our model does not explicitly capture answer information, it still obtains a high answer coverage, which may be because our diversified contexts contain rich answer type words. 

(2) To demonstrate the effectiveness of answer-aware loss, we set the weight of answer-aware loss ($\lambda$) to 0.05/0.2/0.5/1.0 (the last four lines in Table \ref{tab: answer-aware loss}). It can be seen that our model, incorporating answer-aware loss, has a significant improvement on answer coverage while there is no performance degradation on BLEU4 compared with $\lambda=0$, which indicates that answer-aware loss contributes to generating better questions. Especially, the generated questions are more precise because they refer to more definitive answers with high Ans$_{\rm cov}$.

(3) It tends to correspond to alternative answers (object in the triplet fact) for some predicates such as \texttt{fb:location/containedby}, while other predicates (e.g. \texttt{fb:person/gender}) may refer to a definitive answer. To investigate our model, by incorporating answer-aware loss, does not generate an answer type word in a mandatory way,  we found 20.5\% predicate corresponds to the generated questions without answer type words when our model obtains the highest Ans$_{\rm cov}$ ($\lambda$=0.5), and it is very close to 21.7\% for the one in human-annotated questions. This demonstrates that the answer-aware loss does not force all predicates to generate questions with answer type words. 

\subsection{Ablation Study}
\label{subsec:Ablation Study}
\begin{table}[h]\footnotesize
    \centering
    \setlength{\tabcolsep}{4pt}
    \renewcommand\arraystretch{1.1}
    \begin{tabular}{p{3.0cm}<{\centering}p{1.15cm}<{\centering}p{1.15cm}<{\centering}p{1.2cm}<{\centering}}
        \hlinew{1.2pt}
        Model $\quad \ \ $ & BLEU4 & ROUGE$\rm _{L}$ & METEOR \tabularnewline
        \hlinew{1.2pt}
        Our Model$_{\rm ans\_loss}$ & \textbf{41.72} & \textbf{69.31} & \textbf{48.13} \tabularnewline
        \hline
        w/o context copy & 41.27 & 68.36 & 47.54 \tabularnewline
        w/o KB copy & 41.04 & 68.66 & 47.72 \tabularnewline
        w/o answer-aware loss& 41.09 & 68.68 & 47.75 \tabularnewline
        w/o diversified contexts & 40.53 & 68.52 & 47.66 \tabularnewline
        \hlinew{1.2pt}
    \end{tabular}
    \caption{Ablation study by removing the main components, where ``w/o" means without, and ``w/o diversified contexts" represents that diversified contexts are replaced by contexts used in \newcite{N18-1020}.}
    \label{tab: Ablation}
\end{table}
In order to validate the effectiveness of model components, we remove some important components in our model, including context copy, KB copy, answer-aware loss and diversified contexts. The results are shown in Table \ref{tab: Ablation}. We can see that removing any component brings performance decline on all metrics. It demonstrates that all these components are useful. Specifically, the last line in Table \ref{tab: Ablation}, replacing diversified contexts with contexts used in \newcite{N18-1020}, has more obvious performance degradation.

\subsection{Performances on Naturalness} 
\label{subsec:Human Evaluation}
\begin{table}[h]\footnotesize
    \centering
    \setlength{\tabcolsep}{4pt}
    \renewcommand\arraystretch{1.1}
    \begin{tabular}{p{2.8cm}<{\centering}p{2.5cm}<{\centering}}
        \hlinew{1.2pt}
        Model & Naturalness \tabularnewline
        \hlinew{1.2pt}
        \newcite{P16-1056} & 2.96 \tabularnewline
        \newcite{N18-1020} & 2.23 \tabularnewline
        Our Model$_{\rm ans\_loss}$ & \textbf{3.56} \tabularnewline
        \hlinew{1.2pt}
    \end{tabular}
    \caption{Performances on naturalness.}
    \label{tab: naturalness}
\end{table}

Human evaluation is important for generated questions. Following \newcite{N18-1020}, we sample 100 questions from each system, and then two annotators measure the naturalness by a score of 0-5. The Kappa coefficient for inter-annotator is 0.629, and p-value for all scores is less than 0.005. As shown in Table \ref{tab: naturalness}, \newcite{N18-1020} perform poorly on naturalness, while our model obtains the highest score on naturalness, which demonstrates our model can deliver more natural questions than baselines.

\subsection{Discussion}

\subsubsection{Without Pre-trained KB Embeddings} 
\label{subsec:Pre-trained KB}
\begin{table}[h]\footnotesize
    \centering
    \setlength{\tabcolsep}{4pt}
    \renewcommand\arraystretch{1.1}
    \setlength\tabcolsep{0.35pt} 
    \begin{tabular}{p{2.58cm}<{\centering}p{1.0cm}<{\centering}p{1.13cm}<{\centering}p{1.33cm}<{\centering}p{1.36cm}<{\centering}}
        \hlinew{1.2pt}
        Model& TransE & BLEU4 & ROUGE$\rm _{L}$ & METEOR\tabularnewline
        \hlinew{1.2pt}
        \newcite{N18-1020} & True & 36.56 & 58.09 & 34.41 \tabularnewline
        \newcite{N18-1020} & False & 33.67 & 55.57 & 33.20 \tabularnewline
        \hline
        Our Model$_{\rm ans\_loss}$ & True & 41.72 & 69.31 & 48.13 \tabularnewline
        Our Model$_{\rm ans\_loss}$ & False & 41.55 & 68.59 & 47.52 \tabularnewline
        \hlinew{1.2pt}
    \end{tabular}
    \caption{Performances of whether using the pre-trained KB embedding by transE.}
    \label{tab: pre-train}
\end{table}
Pre-trained KB embeddings may provide rich structured relational information among entities. However, it heavily relies on large-scale triplets, which is time and resource-intensive. To investigate the effectiveness of pre-trained KB embedding for KBQG, we report the performance of KBQG whether using pre-trained KB embeddings by simply applying TransE. Table \ref{tab: pre-train} shows that the performance of KBQG is degraded without TransE embeddings. In comparison, \newcite{N18-1020} obtain obvious degradation on all metrics while there is only a slight decline in our model. We believe that it may owe to the context-augmented fact encoder since our model drops to 40.87 on the BLEU4 score without context-augmented fact encoder and transE embeddings.

\subsubsection{The Effectiveness of Generated Questions for Enhancing Question Answering over Knowledge Bases} 
\label{sec:Question Answering}

\begin{table}[h]\footnotesize
    \centering
    \setlength{\tabcolsep}{4pt}
    \renewcommand\arraystretch{1.1}
    \begin{tabular}{p{4.1cm}<{\centering}p{1.5cm}<{\centering}}
        \hlinew{1.2pt}
        Data Type & Accuracy \tabularnewline
        \hlinew{1.2pt}
        human-labeled data & 68.97 \tabularnewline
        \hline
        
        + gen\_data  (Serban et al., 2016) & 68.53 \tabularnewline
        + gen\_data  (Elsahar et al., 2018) & 69.13 \tabularnewline
        + gen\_data (Our Model$_{\rm ans\_loss}$) & \textbf{69.57} \tabularnewline
        \hlinew{1.2pt}
    \end{tabular}
    \caption{Performances of generated questions for QA.}
    \label{tab: Question Answering}
\end{table}

Previous experiments demonstrate that our model can deliver more precise questions. To further prove the effectiveness of our model, we will see how useful the generated questions are for training a question answering system over knowledge bases. Specifically, we combine human-labeled data with the same amount of model-generated data 
to a typical QA system (Mohammed et al., 2018). The accuracy of QA is shown in Table \ref{tab: Question Answering}. We can observe that adding generative questions may weaken the performance of QA (drop from 68.97 to 68.53 in Table \ref{tab: Question Answering}). Our generated questions achieve the best performance on the QA system. It indicates that our model generates more precise question and has improved QA performances greatly. 

\subsubsection{Speed}
\label{subsec:Speed}
\begin{figure}[h]
	\begin{center}
		\includegraphics[width=6.6cm]{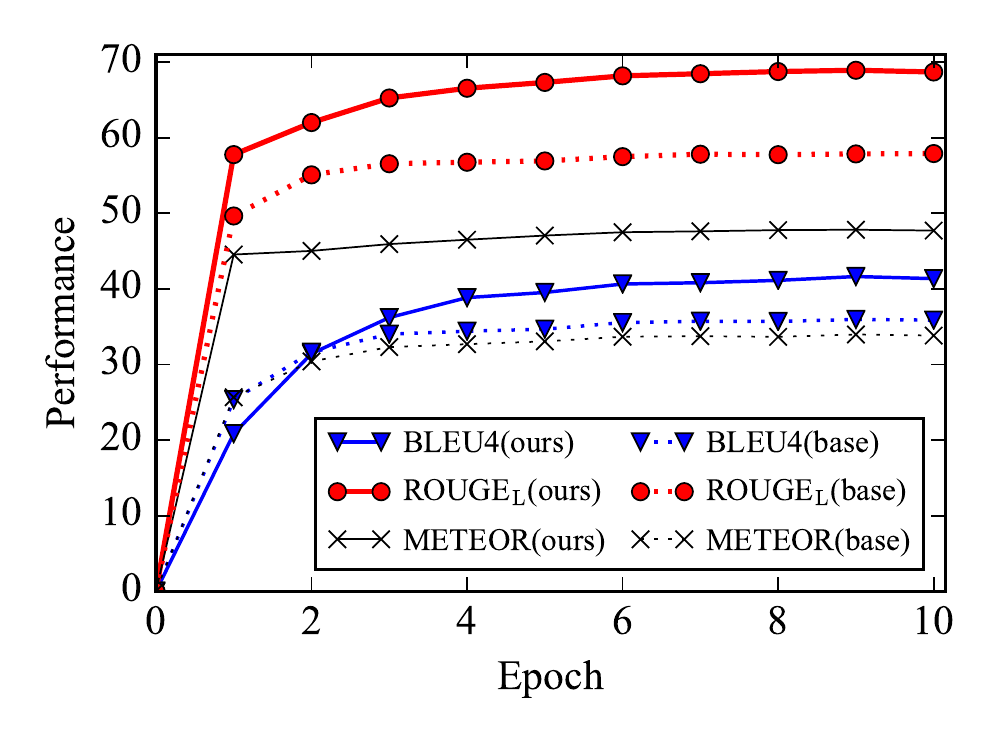}
		\caption{Performance on valid data through epochs, where ``base" is the method in Elsahar et al. (2018).}
		\label{fig:epoch}
	\end{center}
\end{figure}
In order to further explore the convergence speed, we plot the performances on valid data through epochs in Figure \ref{fig:epoch}. Our model has much more information to learn, and it may have a bad impact on the convergence speed. Nevertheless, our model can copy KB elements and textual context simultaneously, which may accelerate the convergence speed. As demonstrated in Figure \ref{fig:epoch}, our model achieves the best performances on almost epochs. After about 6 epochs, performances on our model become stable and convergent. 

\subsubsection{Case Study}
\label{subsec:case_study}
\begin{figure}[h]
    \begin{center}
        \includegraphics[width=7cm]{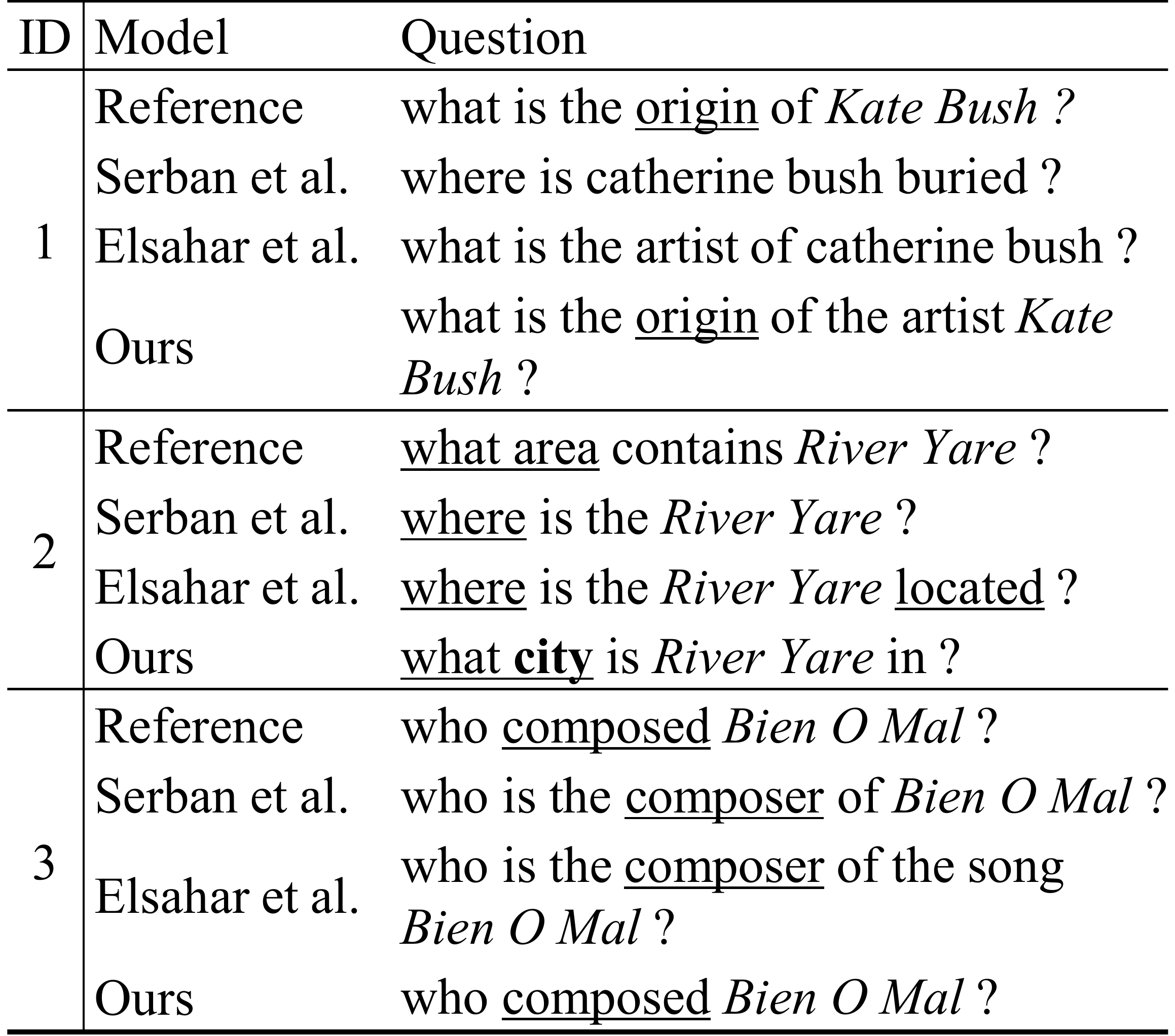}
        \caption{Examples of questions by different models.}
        \label{fig:case}
    \end{center}
\end{figure}
Figure \ref{fig:case} lists referenced question and generated questions by different models. It can be seen that our generated questions can better express the target predicate such as ID 1 (marked as \underline{underline}). In ID 2, although all questions express the target predicate correctly, only our question refers to a definitive answer since it contains an answer type word ``city" (marked as \textbf{bold}). It should be emphasized that the questions, generated by our method with answer-aware loss, do not always contain answer type words (ID 1 and 3).

\section{Related Work}
Our work is inspired by a large number of successful applications using neural encoder-decoder frameworks on NLP tasks such as machine translation \cite{conf/ssst/ChoMBB14} and dialog generation \cite{journals/corr/VinyalsL15}.
Our work is also inspired by the recent work for KBQG based on encoder-decoder frameworks. \newcite{P16-1056} first proposed a neural network for mapping KB facts into natural language questions. To improve the generalization, \newcite{N18-1020} introduced extra contexts for the input fact, which achieved significant performances. However, these contexts may make it difficult to generate questions that express the given predicate and associate with a definitive answer. Therefore, we focus on the two research issues: expressing the given predicate and referring to a definitive answer for generated questions.

Moreover, our work also borrows the idea from copy mechanisms. Point network \cite{NIPS2015_5866} predicted the output sequence directly from the input, and it can not generate new words while CopyNet \cite{P16-1154} combined copying and generating. \newcite{DBLP:conf/aaai/BaoTDYLZZ18} proposed to copy elements in the table (KB). \newcite{N18-1020} exploited POS copy action to better capture textual contexts. To incorporate advantages from above copy mechanisms, we introduce KB copy and context copy which can copy KB element and textual context, and they do not rely on POS tagging.

\section{Conclusion and Future Work}
In this paper, we focus on two crucial research issues for the task of question generation over knowledge bases: generating questions that express the given predicate and refer to definitive answers rather than alternative answers. For this purpose, we present a neural encoder-decoder model which integrates diversified off-the-shelf contexts and multi-level copy mechanisms. Moreover, we design an answer-aware loss to generate questions that refer to definitive answers. Experiments show that our model achieves state-of-the-art performance on automatic and manual evaluations.

For future work, we investigate error cases by analyzing the error distributions of 100 examples. We find that most generated questions (51\%) are judged by the human to correctly express the input facts, but they unfortunately obtain low scores on the widely used metrics. It implies that it is still intractable to evaluate generated questions. Although we additionally evaluate on predicate identification and answer coverage, these metrics may be coarse and deserve further study.

\section*{Acknowledgments}
This work is supported by the National Natural Science Foundation of China (No.61533018), the Natural Key R\&D Program of China (No.2018YFC0830101), the National Natural Science Foundation of China (No.61702512) and the independent research project of National Laboratory of Pattern Recognition. This work was also supported by CCF-Tencent Open Research Fund.
\bibliography{emnlp-2019}
\bibliographystyle{acl_natbib}

\end{document}